\documentclass{svproc}
\usepackage[utf8]{inputenc}
\usepackage{color}
\usepackage{graphicx}

\usepackage{amsmath}
\DeclareMathOperator{\Tr}{Tr}

\usepackage{tikz}
\tikzset{font={\fontsize{9pt}{12}\selectfont}}
\usetikzlibrary{shapes.geometric, arrows}
\tikzstyle{startstop} = [rectangle, rounded corners, minimum width=2cm, minimum height=0.7cm,text centered, draw=black, fill=red!30]
\tikzstyle{stop} = [rectangle, rounded corners, minimum width=2cm, minimum height=1.5cm,text centered, draw=black, fill=red!30]
\tikzstyle{arrow} = [thick,->,>=stealth]
\tikzstyle{process} = [rectangle, minimum width=2cm, minimum height=0.7cm, text centered, draw=black, fill=orange!30]
\usetikzlibrary{positioning}

\usepackage{makeidx}  

\makeindex
\usepackage{url}

\begin{document}
\frontmatter          
\pagestyle{headings}  
%
%
\mainmatter              
\title{Dynamics Based Features for Graphs Classification}
\titlerunning{Dynamics Based Features for Graphs Classification}  
%
\author{Leonardo Gutiérrez Gómez \inst{1,3}, Benjamin Chiêm \inst{1,4} \and  Jean-Charles Delvenne \inst{1,2,5} }
\authorrunning{Gutierrez et al.} 
%

%
\institute{Université catholique de Louvain, Louvain-la-Neuve, Belgium,\\
Institute of Information and Communication Technologies\\
Electronics and Applied Mathematics (ICTEAM)
\and Centre for Operations Research and Econometrics (CORE)
\and
\email{leonardo.gutierrez@uclouvain.be},
\and
\email{benjamin.chiem@student.uclouvain.be}
\and 
\email{jean-charles.delvenne@uclouvain.be}
}

\maketitle              

\begin{abstract}
Numerous social, medical, engineering and biological challenges can be framed as graph-based learning tasks. Here, we propose a new feature based approach to network classification. We show how dynamics on a network can be useful to reveal patterns  about the organization of the components of the underlying graph where the process takes place. We define generalized assortativities on networks and use them as generalized features across multiple time scales. These features turn out to be suitable signatures for discriminating between different classes of networks.
Our method is evaluated empirically on established network benchmarks. We also introduce a new dataset of human brain networks (connectomes) and use it to evaluate our method. Results reveal that our dynamics based features are competitive and often outperform state of the art accuracies.
\end{abstract}
\section{Introduction}

A wide range of real world problems involve network analysis and prediction tasks. The complexity of social, engineering and biological networks makes necessary developing methods to deal with complex challenges arisen by mining graph-based data. A common issue that occurs often in applications is determining differences or similarities between graphs, identifying common connectivity patterns or nodes features which are informative enough for recognition purposes.

In a typical graph classification task, we are interested in assigning the most likely label to a graph among a set of classes. For instance, in chemoinformatics, bioinformatics, one is interested in predicting the toxicity or anti-cancer activity of molecules. Characterization of proteins and enzymes is crucial in drugs design in order to discover the apparition of diseases or defective compounds. Social networks classification \cite{Wang2012} is suitable for many marketing, political or targeting purposes, as well as mobility, collaboration networks and so forth. 

This problem has been addressed before from the supervised and unsupervised machine learning perspective. In the first case, a set of discriminative hand-crafted features or descriptors must be carefully chosen in order to enhance generalization capabilities. Typically it is done by manually choosing a set of structural, global and contextual features \cite{Fei:2008:SFS:1458082.1458212} from the underlying graph. Kernel-based methods \cite{Hofmann2008} are well known in the literature. Roughly speaking they define a similarity measure between two objects as an inner product in a reproducing kernel Hilbert space (RKHS) assessed through a \textit{kernel function}. A classification step is performed afterwards using a kernelized form of a classification algorithm such as Support Vector Machines. However, it is always challenging to come up with a rich kernel that keeps the problem computationally tractable. Alternatively, unsupervised algorithms aim to learn those features from data typically through a neural network variant. Meanwhile, the high training cost and often huge amount of parameters makes this alternative intractable for large complex networks.

In this paper we build on the idea that the dynamics of a random walker on a network is strongly indicative of the structure of the network in which it takes place, an idea used e.g. in community detection \cite{Delvenne2013} or node centrality \cite{Brin1998}, and use features of the dynamics of the random walker as network descriptors. Specifically, those network features are obtained in a systematic way from the correlation patterns of node attributes seen by a random walker at different time instants. These correlations are seen as generalized assortativities of the nodes attributes over the whole network. The attribute can be structural (degree, Pagerank etc.) or exogenous labels (e.g. atom type in a molecule network, age in a social network). These descriptors are then computed over many bioinformatic, social and neuroscience graph datasets and therefore used to train a classification model in order to perform binary and multi-class network classification tasks. We compare experimentally its classification accuracy with respect to some representative graph kernels and features based algorithms of the literature. 

Our results shown that that generalized assortativities on small set of elementary structural patterns, and any available exogeneous node label if available, are capable of achieving and in many cases, outperform state-of-the-art accuracies, being simple, efficient and versatile in benchmark network datasets. 
Finally, we introduce a new human brains dataset we built from MRI data, and use it to evaluate our method.

The paper is structured as follows: Section 2 reviews similar approaches in the literature. Section 3 introduces generalized assortativities. Section 4 addresses the problem of definition of features of network dynamics. Finally, in sections 5 and 6 we present our experimental setting in which we assessed our method and discussion.
\section{Related work}
Graph classification has been extensively studied by the networks/machine learning communities under different headings. Methods can be grouped in three categories: feature based models, graph kernels and neural networks approaches. 

There is a considerable amount of literature related with graph kernel-based methods. They can be categorized in three classes: graph kernels based in \textit{random walks} \cite{Borgwardt}, \cite{DBLP:conf/icml/KashimaTI03}, kernels based on \textit{subtree patterns}, \cite{Ramon03expressivityversus}, \cite{Shervashidze2009FastSK}, \cite{Shervashidze:2011:WGK:1953048.2078187},  and also kernels based on \textit{limited-size subgraphs} or \textit{graphlets}, \cite{5664}, \cite{Horvath2005}. The similarity between graphs is assessed by counting the number of common substructures, or decomposing the input graphs in substructures such as shortest path or neighborhood subparts. However, the aforementioned graph kernels scale poorly for relatively large networks. When the number of structures is exponential or infinite in the case of random walks perspective, the kernel between two graphs is computed without generating an explicit feature mapping. Thus, the complexity grows quadratically in the number of examples  because the memory needed to compute the kernel matrix. This issue remains as an unsolved challenge for large scale networks.  

Improved graph kernels such that Weisfeiler-Lehman (WL) Subtree Kernel \cite{Shervashidze2009FastSK}, and Neigborhood Subgraph Pairwise Distance Kernel \cite{267297} scale better by choosing a pattern that generate explicitly the feature map in polynomial space complexity. A recent work \cite{Yanardag:2015:DGK:2783258.2783417} proposes a deep version of the Graphlet, WL and Shortest-Path kernels. Patterns are generated with a CBOW or Skip-gram model in the same way that words are embedded in a Euclidean space in a natural language processing model \cite{journals/corr/abs-1301-3781}.\\
Automatic feature learning algorithms aim to learn the underlying graph patterns often through a neural network variant. More recently \cite{Niepert2016} proposes to learn a convolution neural network (PSCN) for arbitrary graphs, by exploiting locally connected regions directly from the input graph. Shift aggregate extract network  (SAEN) \cite{DBLP:journals/corr/OrsiniBF17} introduces a neural network to learn graph representations, by decomposing the input graph hierarchically in compressed objects, exploiting symmetries and reducing memory usage.\\
Barnett \cite{Barnett2016} proposes an hybrid feature-based approach. It combines manual selection of network features with existing automated classification methods. \\
Our method fits the feature-based models. Feature vectors are created through a dynamic taking place on the network. This dynamic aims to capture correlations along different network attributes and external metadata. Our method is general and versatile being flexible to work with a wide range of heterogeneous graphs.

\section{Random walks and generalized assortativities}



Consider an undirected, connected and non-bipartite graph $G=(V,E)$ with $N$ vertices and $m$ edges. We will assume the graph is unweighted, but all the results are applicable to the weighted case (see section \ref{humanbds}). Extension to directed networks is straightforward as well. The adjacency matrix $A$ associated to $G$ is an $N \times N$ binary matrix, with $a_{ij}=1$ if the vertices $i$ and $j$ are connected, and $0$ otherwise. The number of edges in the vertex $i$ is known as the \textit{degree} of the vertex $i$, denoted $d_i$. Then the degree vector of $G$ can be compiled as $\textbf{d} = A \textbf{1}$, where \textbf{1} is a $N \times 1$ ones vector. We will also define the $N \times N$ degree matrix, $D=diag(d)$.

The standard random walk on such a graph defines a Markov chain in which transition probabilities are split uniformly among the edges, with a transition probability $1/d_i$:

\begin{equation}\label{markov}
\textbf{p}_{t+1} = \textbf{p}_t [D^{-1}A ] \equiv \textbf{p}_t M
\end{equation}

Here $\textbf{p}_t$ is the $1 \times N$ normalized probability distribution of the process at time $t$, and $M$ the $N \times N$ transition matrix. Under the assumptions on the graph (connected, undirected, and non-bipartite), the dynamics converges to a unique \textit{stationary distribution} vector $\pi = d^T/2m$.
We define also the $N \times N$ diagonal matrix $\Pi = diag(\pi)$. 

Consider a scalar node attribute such as the degree, centrality, age, etc. It is known that many real networks exhibit interesting assortativity properties, i.e. a high covariance between neighbor's attributes of a randomly selected edge. For the random walker in stationary distribution, this translates into the covariance  
of the node attribute visited at two consecutive time instants $\tau$ and $\tau+1$, since the transition can occur on each edge of the graph with identical probability. 

This suggests to consider generalized assortativities, as the covariance between the attributes seen by the walker at times $\tau$ and $\tau + t$, for any $t=0,1,2,3,\ldots$. While the case $t=0$ is simply the variance, the case $t>1$ explores systematically assortativity patterns beyond the direct neighborhood. \\

Let us define the indicator variable, $X_k(t)=1$ if the walker is at node $k$ at time $t$. $X_t = [X_1(t), X_2(t), \ldots, X_N(t)]$ is an indicator vector and its \textit{autocovariance} matrix at time $t$  is expressed as
\begin{equation}\label{covariance}
\rho (t) = Cov(X_{\tau}, X_{\tau + t}) = \Pi M^t - \pi^T \pi
\end{equation}

Given  an $N \times 1$ node attribute vector $v$, its attribute covariance is given by

\begin{equation}\label{vector_covariance}
u_v(t) =  v^T \rho(t) v
\end{equation}
It often happens that a node attribute is categorical rather than numerical, e.g. gender or atom type. In this case one can still compute an assortativity coefficient by encoding each category as a binary characteristic vector. One therefore encode  
a $k$-category attribute as an $N \times k$ binary matrix $H$, and compute the covariance of each category, i.e. each column $h_i$ of $H$. In order to get a single generalized assortativity coefficient, one sums over all categories and obtain

\begin{equation}\label{autocovarience}
r(t,H) = \Tr \left[ H^T\rho(t) H \right] 
\end{equation}

It can be noted that a categorical  $H$ can be seen as defining a partition of the nodes, and $r(1,H)$ happens to coincide with the modularity of the partition \cite{Newman2006,Delvenne2013} used in community detection, or graph clustering. The quantity $r(t,H)$, called in this context \emph{Markov stability} \cite{Delvenne2013}, allows multi-scale community detection by scanning through the time parameter $t$.

\begin{figure}[!ht]\label{stabexample}
\begin{center}
\includegraphics[width=3.0in]{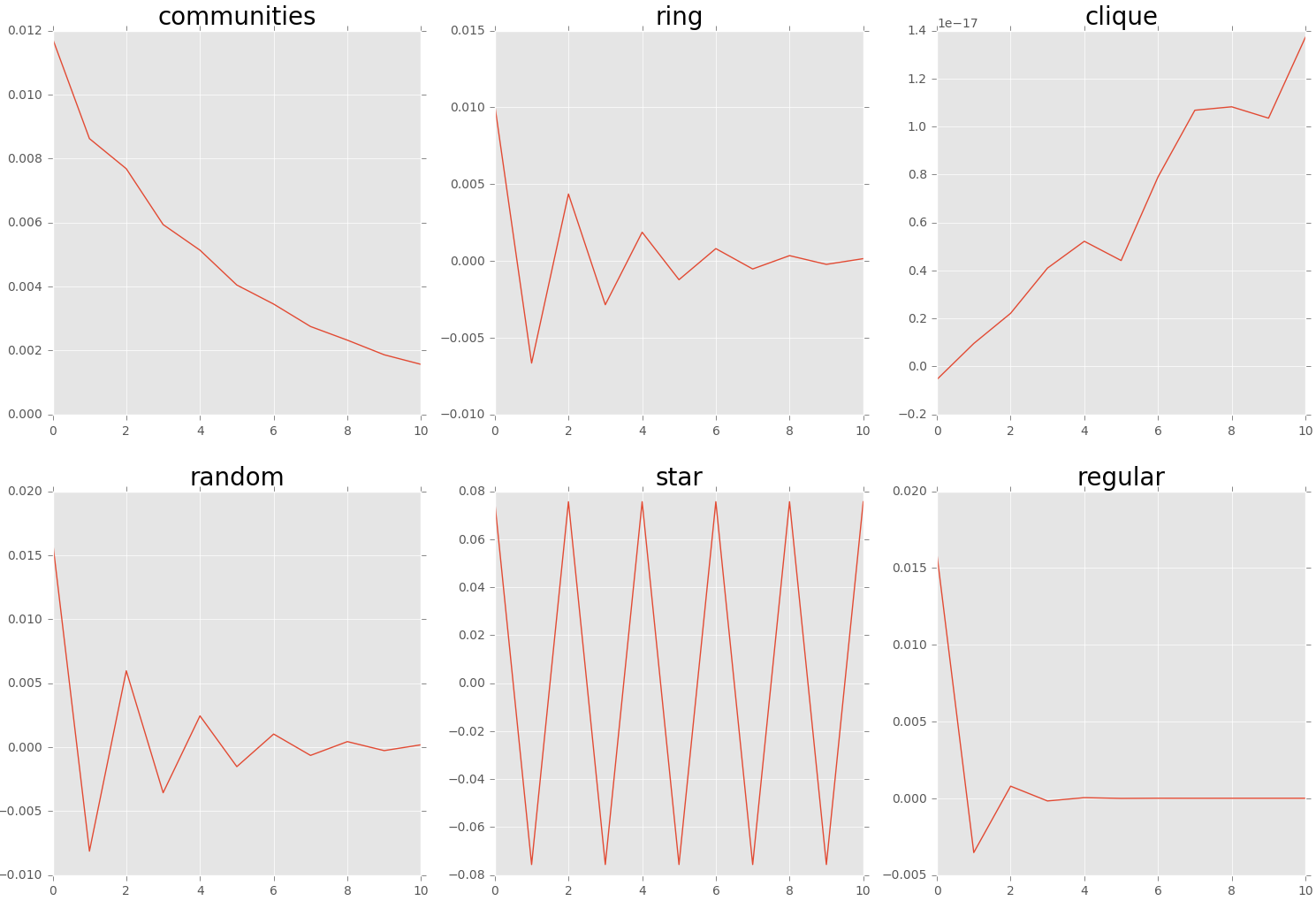}
\caption{Generalized assortativity vs time with respect to second left eigenvector of transition matrix on six different network topologies: (1) network with 3 well defined communities, (2) ring network, (3) clique, (4) Erdős-Rényi random graph with $p=0.4$, (5) star (6) regular graph} 
\label{stab_secondeigenvector}
\end{center}
\end{figure}
\section{Features of network dynamics}
One take the simplest structural node attributes, such as the degree (or equivalently, PageRank, which is the dominant left eigenvector of $M$), the second left eigenvector of $M$,  the local clustering coefficient, for different times $t=0,1,2,3, \ldots$ and generate network features. 
  Categorical node metadata, whenever available (i.e atom type or protein labels, in bioinformatic graphs), can be added to produce even more features.

The simplest node attribute is certainly the node ID  itself, i.e. $H=I$ (identity matrix) in (\ref{autocovarience}). For $t=0$ it yields an assortativity $r(0,I)=1- \|\pi\|^2$, delivering effectively a variance of the degree distribution. 
For $t=3$, it counts essentially the probability for the random walker to follow a triangle, thus representing a variant of the global clustering coefficient.


It turns out that this attribute covariances are useful to distinguish between network classes. They are naturally dataset dependent, meaning that features that discriminate well in some networks could be less performance in other networks (i.e node clustering coefficient in molecules vs social networks). 

This hypothesis is illustrated with a toy example in figure (\ref{stab_secondeigenvector}). Plots correspond to the second eigenvector covariances for $t \in [0,10]$. It can be seen how $u_v(t)$ varies among different graph topologies. It is worth to mention that the most useful information is concentrated at the beginning of the process. Because the covariance decreases relatively quickly to zero, large values of $t$ should give useless information. Therefore our computations were done with $t \in \{0,1,2,3 \}$. 

All or some of the generalized assortativities described above are then compiled in a feature vector and can be combined with other global graph attributes i.e. number of nodes/edges. This vector is used as a discriminative signature of the network for classification purposes.

\section*{Experiments}
The aim of our experiments is threefold. First, we want to show that a small set of generalized assortativity features is enough to obtain comparable results and often outperform state of the art algorithms in terms of classification accuracy on graph benchmark datasets. Second, we test the versatility of our method by applying it on networks with heterogeneous properties, nodes attributes, weighted edges and so forth. Finally we show how our method can be optimized on datasets in which all graphs have the same set of nodes outperforming considerably algorithms of the literature.\\
We consider two cases in our experimental setup differing only on the datasets we used. The former (Benchmark experiments) corresponds to the application of our method on bioinformatic/social networks datasets. In the second one we create a new human brains dataset and use it to evaluate our method.\\ 

We setup a general set of node attributes as possible candidates for the feature selection process. Indeed, we consider the identity partition, .i.e $H=I$ as the base-line feature. First and second dominant left eigenvectors of the transition matrix are also considered. Local clustering coefficient, node betweeness and number of triangles per node are included as well as categorical node labels. Number of nodes and edges are considered also, as the unique global features of our model.\\
According with the dataset, we chose the minimal subset of attributes in which our assortativity features gave us the best classification accuracies.
\section{Benchmark experiments}
\subsection{Datasets}

A dozen of real-world datasets were used in our experiments. We summarize their global properties in Table  \ref{datasets}. The first six belong to the bioinformatic class, composed by undirected graphs representing either molecules, enzymes and proteins. They all contain categorical node labels encoding a particular functional property related with an specific task. The later six are unlabeled node social network datasets. 

\textbf{MUTAG} is a nitro compounds dataset divided in two classes according to their mutageic activity on bacterium \textit{Salmonella Typhimurium}. \textbf{PTC} dataset contains compounds labeled according to carcinogenicity 
on rodents, divided in two groups. Vertices are labeled by 20 atom types. The \textbf{NC11} and \textbf{NCI109} from the National Cancer Institute (NCI), are two balanced dataset of chemical compounds  screened for activity against non-small cell lung cancer and ovarian cancer cell. They have 37 and 38 categorical  node labels respectively. \textbf{PROTEINS} is a two classes dataset in which nodes are secondary structure elements (SSEs). Nodes are connected if they are contiguous in the aminoacid sequence. \textbf{ENZYMES} is a dataset of protein tertiary structures consisting of 600 enzymes from the BRENDA enzyme database. The task is to assign each enzyme to one of the 6 EC top-level classes.  
\begin{table}[!h]
\scriptsize
\caption{Statistics for benchmark datasets. First six correspond to bioinformatic and the rest to social network datasets. NA. meaning unlabeled nodes}
\begin{center} \label{datasets}
\begin{tabular}{l@{\quad}c@{\quad}c@{\quad}c@{\quad}c@{\quad}c}
\hline
Dataset & Num graphs & Classes & Node labels & Avg nodes & Avg edges \\ 
\hline\rule{0pt}{12pt} 
MUTAG & 188 & 2 & 7 & 17.93 & 19.79 \\ [2pt]
 
PTC & 344 & 2 & 19 & 14.29 & 14.69 \\ [2pt]
 
NCI1 & 4110 & 2 & 37 & 29.87 & 32.3 \\ [2pt]
 
NCI109 & 4127 & 2 & 38 & 29.68 & 32.13 \\ [2pt]

ENZYMES & 600 & 6 & 3 & 32.63 & 62.14 \\ [2pt]

PROTEINS & 1113 & 2 & 3 & 39.06 & 72.82 \\ [2pt]
\hline\rule{0pt}{12pt} 
COLLAB & 5000 & 3 & NA & 74.49 & 2457.78 \\ [2pt]

REDDIT-BINARY & 2000 & 2 & NA & 429.63 & 497.75 \\[2pt] 

REDDIT-MULTI-5K & 5000 & 5 & NA & 508.52 & 594.87 \\ [2pt]

REDDIT-MULTI-12K & 11929 & 11 & NA & 391.41 & 456.89 \\ [2pt]

IMDB-BINARY & 1000 & 2 & NA & 19.77 & 96.53 \\ [2pt]

IMDB-MULTI & 1500 & 3 & NA & 13.0 & 65.94 \\ [2pt]
\hline\rule{0pt}{12pt} 
\end{tabular} 
\end{center}
\end{table}
From the social networks pool, \textbf{COLLAB} is a scientific-collaboration dataset, where ego-networks of researchers that have collaborated together are constructed. The task is then determine whether the ego-collaboration network belongs to any of three classes, namely, \textit{High Energy Physics}, \textit{Condense Mater Physics} and \textit{Astro Physics}. \textbf{REDDIT-BINARY}, \textbf{REDDIT-MULTI-5K} and \textbf{REDDIT-MULTI-12K} are three balanced two, five and eleven classes datasets respectively. Each one contains a set of graphs representing an on-line discussion thread where nodes corresponds to users and there is an edge between them if anyone responds to another's comment. The task is then to  discriminate between threads from which the subreddit was originated. \textbf{IMDB-BINARY} is an ego-networks of actors that have appeared together in any movie.  Graphs are constructed from \textit{Action} and \textit{Romance} genres. The task is identify which genre an ego-network graph belongs to. \textbf{IMDB-MULTI} is the same, but consider three classes: \textit{Comedy}, \textit{Romance} and \textit{Sci-Fi}.


\subsection{Experimental setup}

We assess the performance of our method against some representative graph kernels, feature-based and neural networks  methods of the literature. We choose the Graphlet \cite{Shervashidze:2011:WGK:1953048.2078187}, Shorthest path \cite{Borgwardt} and the Weisfeiler-Lehman subtree kernels \cite{Shervashidze:2011:WGK:1953048.2078187}, as well as their respective deep versions \cite{Yanardag:2015:DGK:2783258.2783417}. Random walks based kernels  as  p-step random walk \cite{Smola2003}, the random walk \cite{Gartner03ongraph} and Ramon \& Gartner kernels \cite{Ramon03expressivityversus} are also considered. We also compare against the feature-based method \cite{Barnett2016}, the convolutional neural network PSCN \cite{Niepert2016} and the Shift aggregate extract network (SAEN) \cite{DBLP:journals/corr/OrsiniBF17}.

For graph-kernels methods we used the Matlab scripts\footnote{http://www.mit.edu/$\sim$pinary/kdd/}. The Deep Graph kernel scripts were written in Python by his author and were taken from his website\footnote{http://www.mit.edu/$\sim$pinary/kdd/}. We coded the feature-based approach \cite{Barnett2016} and our algorithm in Matlab. We make available the source code in this site\footnote{http://sites.uclouvain.be/big-data/Downloads/dynfeats}. \\

In order to make as fair as possible a comparison, we performed 10-fold cross validation of linear C-Support Vector Machine on bioinformatic datasets and Random Forest for social networks. Each dataset is split in 9 folds for training and 1 fold for testing. Parameters C of SVM and number of trees for Random Forest were optimized on the training set only. In order to exclude the random effect of the fold assignment, we repeated this experiment 10 times. We report the \textit{average prediction accuracies} and its \textit{standard deviation}. 

Parameters for the graph kernels are cross-validated  on the training set following \cite{Shervashidze:2011:WGK:1953048.2078187} and \cite{Yanardag:2015:DGK:2783258.2783417} settings:
\begin{itemize}
\item[$\bullet$]  For Weisfeiler-Lehman subtree kernel, $h$ is taken from $\{0,1,2,\ldots,10 \}$
\item[$\bullet$] The $p$ value for $p$-step random walk kernel is chosen from $\{1,2,3 \}$
\item[$\bullet$] We computed the random walk kernel for the decay $\lambda \in \{ 10^{-6}, 10^{-5},\ldots, 10^{-1} \}$
\item[$\bullet$] The height parameter in Ramon \& Gartner kernel is taken from $\{1,2,3 \}$
\item[$\bullet$] For the Deep Graph kernels (GK, SP and WL), the window size and feature dimension is chosen from $\{2,5,10,25,50 \}$
\end{itemize}
For each kernel we report the result for the parameter that achieves the best classification accuracy. For the feature-based approach \cite{Barnett2016}, feature vectors were built with the same network features they reported in their paper: number of nodes, number of edges, average degree, degree assortativity, number of triangles and global clustering coefficient. Finally, for PSCN we compare with the accuracies they reported in \cite{Niepert2016}. Same with SAEN \cite{DBLP:journals/corr/OrsiniBF17}.
\begin{table}[!h] 
\caption{\textit{Graph-kernels methods: } Mean and standard deviation of classification accuracy for Random Walk (RW) \cite{Gartner03ongraph}, Weisfeiler-Lehman (WL) \cite{Shervashidze:2011:WGK:1953048.2078187}, Graphlet (GK)\cite{5664}, Deep Graphlet kernels (DGK) \cite{Yanardag:2015:DGK:2783258.2783417},  Dynamics Based Features (DyF) (our method) }
\begin{center}\label{social1}
\scriptsize
\begin{tabular}{l@{\quad}c@{\quad}c@{\quad}c@{\quad}c@{\quad}c@{\quad}}
\hline  \rule{0pt}{10pt} 
Dataset	&	RW	&	WL	&	GK	&	DGK	&	DyF	\\[2pt]
\hline  \rule{0pt}{10pt} 
COLLAB	&	69.01 $ \pm $ 0.09	&	77.79 $ \pm $0.19	&	72.84 $ \pm $ 0.28	&	73.09 $ \pm $ 0.25	&	\textbf{80.61 $ \pm $ 1.60}	\\[2pt]
IMDB-BINARY	&	64.54 $ \pm $ 1.22	&	72.86 $ \pm $0.76	&	65.87 $ \pm $ 0.98	&	66.96 $ \pm $ 0.56	&	\textbf{72.87 $ \pm $ 4.05}	\\[2pt]
IMDB-MULTI	&	34.54 $ \pm $ 0.76	&	\textbf{50.55 $ \pm $0.55}	&	43.89 $ \pm $ 0.38	&	44.55 $ \pm $ 0.52	&	48.12 $ \pm $ 3.56	\\[2pt]
REDDIT-BINARY	&	67.63 $ \pm $ 1.01	&	69.57 $ \pm $0.88	&	77.34 $ \pm $ 0.18	&	78.04 $ \pm $ 0.39	&	\textbf{89.51 $ \pm $ 1.96}	\\[2pt]
REDDIT-MULTI-5K	&	$>$ 72h	&	47.72 $ \pm $0.48	&	41.01 $ \pm $ 0.17	&	41.27 $ \pm $ 0.18	&	\textbf{50.31 $ \pm $ 1.92}	\\[2pt]
REDDIT-MULTI-12K	&	$>$ 72h	&	38.47 $ \pm $0.12	&	31.82 $ \pm $ 0.08	&	32.22 $ \pm $ 0.10	&	\textbf{40.30 $ \pm $ 1.41}	\\[2pt]
\hline \rule{0pt}{10pt} 
\end{tabular}
\end{center}
\caption{\textit{Neural nets and feature-based approach:} Mean and standard deviation of classification accuracy for Shift agreggate extract network (SAEN) \cite{DBLP:journals/corr/OrsiniBF17}, Convolutional Neural Network (PSCN) \cite{Niepert2016}, Feature-Based (FB) \cite{Barnett2016}, Dynamics Based Features (DyF) (our method) }
\begin{center}\label{social2}
\scriptsize
\begin{tabular}{l@{\quad}c@{\quad}c@{\quad}c@{\quad}c@{\quad}}
\hline  \rule{0pt}{10pt} 
Dataset	&	SAEN	&	PSCN	&	FB	&	DyF	\\
\hline  \rule{0pt}{12pt} 
COLLAB	&	75.63 $ \pm $0.31	&	72.60 $ \pm $ 2.15	&	76.35 $ \pm $ 1.64	&	\textbf{80.61 $ \pm $ 1.60}	\\[2pt]
IMDB-BINARY	&	71.26 $ \pm $ 0.74	&	71.00 $ \pm $ 2.29	&	72.02 $ \pm $ 4.71	&	\textbf{72.87 $ \pm $ 4.05}	\\[2pt]
IMDB-MULTI	&	\textbf{49.11 $ \pm $ 0.64}	&	45.23 $ \pm $ 2.84	&	47.34 $ \pm $3.56	&	48.12 $ \pm $ 3.56	\\[2pt]
REDDIT-BINARY	& 86.08 $ \pm $0.53	&	86.30 $ \pm $ 1.58	&	88.98 $ \pm $ 2.26	&	\textbf{89.51 $ \pm $ 1.96}	\\[2pt]
REDDIT-MULTI-5K	& \textbf{52.24 $ \pm $0.38}	&	49.10 $ \pm $ 0.70	&	50.83 $ \pm $ 1.83	&	50.31 $ \pm $ 1.92	\\[2pt]
REDDIT-MULTI-12K	& \textbf{46.72 $ \pm $ 0.23}	&	41.32 $ \pm $ 0.42	&	42.37 $ \pm $ 1.27	&	40.30 $ \pm $ 1.41	\\[2pt]
\hline \rule{0pt}{10pt} 
\end{tabular}
\end{center}
\end{table}
\subsection{Results}
We test our method on several benchmark datasets and compare our classification accuracies with the ones achieved by the aforementioned algorithms.

We apply on \textbf{social network} datasets those algorithms that can handle unlabeled node graphs. Results for social graphs are depicted in Tables \ref{social1} (\textit{graph-kernels methods}) and \ref{social2} (\textit{neural nets and feature-based approach}).

As can be seen from Table \ref{social1} we outperform all graph-kernels on all social network datasets, except IMDB-MULTI. Regarding Table \ref{social2}, our method perform generally better than Convolutional Neural Nets (PSCN). However, SAEN outperforms our method in IMDB and REDDIT multiclass problems. We also have better accuracies than FB or similarly except for REDDIT-MULTI-12K, see Table \ref{social2}.\\

Results related with \textbf{bioinformatic datasets} are depicted in Tables \ref{bio1-1} (\textit{Methods that do not exploit node labels}), \ref{bio1-2} (\textit{random-walks and Ramon $\&$ Gartner kernel}), and \ref{bio1-3} (\textit{others graph-kernels and neural nets approaches})

\begin{table}[!t]
\caption{\textit{Methods that do not exploit node labels: }Mean and standard deviation classification accuracy for: Graphlet (GK) \cite{5664}, Deep Graphlet (DGK)\cite{Yanardag:2015:DGK:2783258.2783417}, Feature-Based (FB) \cite{Barnett2016}) and our Dynamics Based Features,  without using node labels (DyF-nolab) and with labels (DyF) } 
\begin{center}\label{bio1-1}
\scriptsize
\begin{tabular}{l@{\quad}c@{\quad}c@{\quad}c@{\quad}c@{\quad}c}
\hline \rule{0pt}{12pt} 
Data	&	GK	&	DGK	&	FB	&	DyF-nolab	&	DyF	\\[2pt]
\hline \rule{0pt}{12pt} 
MUTAG	&	81.66 $ \pm $ 2.11	&	82.66 $ \pm $ 1.45	&	84.66 $ \pm $ 2.01	&	\textbf{86.27 $ \pm $  1.33}	&	\textbf{88.00 $ \pm $ 2.37}	\\[2pt]
PTC	&	57.26 $ \pm $ 1.41	&	\textbf{57.32 $ \pm $1.13}	&	55.58 $ \pm $ 2.30	&	56.15 $ \pm $  1.06	&	57.15 $ \pm $ 1.47	\\[2pt]
NCI1	&	62.28 $ \pm $ 0.29	&	62.48 $ \pm $ 0.25	&	62.90 $ \pm $ 0.96	&	\textbf{66.55 $ \pm $  0.26}	&	\textbf{68.27 $ \pm $ 0.34}	\\[2pt]
NCI109	&	62.60 $ \pm $ 0.19	&	62.69 $ \pm $ 0.23	&	62.43 $ \pm $ 1.13	&	\textbf{66.33 $ \pm $  0.22}	&	\textbf{66.72 $ \pm $ 0.20}	\\[2pt]
PROTEINS	&	71.67 $ \pm $ 0.55	&	71.68$ \pm $ 0.50	&	69.97 $ \pm $ 1.34	&	\textbf{73.14 $ \pm $  0.40}	&	\textbf{75.04 $ \pm $ 0.65}	\\[2pt]
ENZYMES	&	26.61 $ \pm $ 0.99	&	27.08 $ \pm $ 0.79	&	\textbf{29.00 $ \pm $ 1.16}	&	26.55 $ \pm $  1.23	&	\textbf{33.21 $ \pm $ 1.20}	\\[2pt]
\hline
\end{tabular}
\end{center}
\caption{\textit{Ramdom-walks and Ramon $\&$ Gartner kernels: }Mean and standard deviation classification accuracy for: Ramon \& Gartner (RG) \cite{Ramon03expressivityversus}, p-Random Walk (pRW)  \cite{Smola2003}, Random Walk (RW) \cite{Gartner03ongraph}) and our Dynamics Based Features,  without using node labels (DyF-nolab) and with labels (DyF) }

\begin{center}\label{bio1-2}
\scriptsize
\begin{tabular}{l@{\quad}c@{\quad}c@{\quad}c@{\quad}c@{\quad}c} 
\hline \rule{0pt}{12pt} 
Data	&	RG	&	pRW	&	RW	&	DyF-nolab	&	DyF	\\[2pt]
\hline \rule{0pt}{12pt} 
MUTAG	&	84.88 $ \pm $  1.86	&	80.05 $ \pm $  1.64	&	83.72 $ \pm $  1.50	&	\textbf{86.27 $ \pm $  1.33}	&	\textbf{88.00 $ \pm $  2.37}	\\[2pt]
PTC	&	58.47 $ \pm $  0.90	&	\textbf{59.38 $ \pm $  1.66}	&	57.85 $ \pm $  1.30	&	56.15 $ \pm $  1.06	&	57.15 $ \pm $  1.47	\\[2pt]
NCI1	&	56.61 $ \pm $  0.53	&	$>$ 72h	&	48.15 $ \pm $  0.50	&	\textbf{66.55 $ \pm $  0.26}	&	\textbf{68.27 $ \pm $  0.34}	\\[2pt]
NCI109	&	54.62 $ \pm $  0.23	&	$>$ 72h	&	49.75 $ \pm $  0.60	&	\textbf{66.33 $ \pm $  0.22}	&	\textbf{66.72 $ \pm $  0.20}	\\[2pt]
PROTEINS	&	70.73 $ \pm $  0.35	&	71.16 $ \pm $  0.35	&	74.22 $ \pm $  0.42	&	73.14 $ \pm $  0.40	&	\textbf{75.04 $ \pm $  0.65}	\\[2pt]
ENZYMES	&	16.96 $ \pm $  1.46	&	30.01 $ \pm $  1.01	&	24.16 $ \pm $  1.64	&	26.55 $ \pm $  1.23	&	\textbf{33.21 $ \pm $  1.20}	\\[2pt]
\hline
\end{tabular}
\end{center}

\caption{\textit{Others graph-kernels and neural nets approaches: }Mean and standard deviation classification accuracy for: Deep Shortest-path (DSP) \cite{Yanardag:2015:DGK:2783258.2783417}, Weisfeiler-Lehman (WL) \cite{Shervashidze:2011:WGK:1953048.2078187}, Deep Weisfeiler-Lehman (DWL) \cite{Yanardag:2015:DGK:2783258.2783417} kernels, Convolutional Neural Network (PSCN) \cite{Niepert2016}) and Dynamics Based Features (DyF) (our method)}
\begin{center}\label{bio1-3}
\scriptsize
\begin{tabular}{l@{\quad}c@{\quad}c@{\quad}c@{\quad}c@{\quad}c} 
\hline \rule{0pt}{12pt} 
Data	&	DSP	&	WL	&	DWL	&	PSCN	&	DyF	\\[2pt]
\hline \rule{0pt}{12pt} 
MUTAG	&	87.44 $ \pm $ 2.72	&	80.72 $ \pm $ 3.00	&	82.94 $ \pm $ 2.68	&	\textbf{92.63 $ \pm $ 4.21}	&	88.00 $ \pm $ 2.37	\\[2pt]
PTC	&	59.52 $ \pm $ 2.19	&	56.97 $ \pm $ 2.01	&	59.17 $ \pm $ 1.56	&	\textbf{60.00 $ \pm $ 4.82}	&	57.15 $ \pm $ 1.47	\\[2pt]
NCI1	&	73.55 $ \pm $ 0.51	&	80.13 $ \pm $ 0.50	&	80.31 $ \pm $ 46	&	78.59 $ \pm $ 1.89	&	68.27 $ \pm $ 0.34	\\[2pt]
NCI109	&	73.26 $ \pm $ 0.26	&	80.22 $ \pm $ 0.34	&	\textbf{80.32 $ \pm $ 0.33}	&	--	&	66.72 $ \pm $ 0.20	\\[2pt]
PROTEINS	&	75.78 $ \pm $ 0.54	&	72.92 $ \pm $ 0.56	&	73.30 $ \pm $ 0.82	&	\textbf{75.89 $ \pm $ 2.76}	&	75.04 $ \pm $ 0.65	\\[2pt]
ENZYMES	&	41.65 $ \pm $ 1.57	&	53.15 $ \pm $ 1.14	&	\textbf{53.43 $ \pm $ 0.91}	&	--	&	33.21 $ \pm $ 1.20	\\[2pt]
\hline
\end{tabular}
\end{center}
\end{table}
In Table \ref{bio1-1} we observe that ignoring node labels and solely based on the structure of the graphs, we generally out-perform GK, DGK and FB which also ignore labels. Similarly, our method exhibits better performances than random-walk based methods as well as RG kernel even in the case we do not take labels into account, see Table \ref{bio1-2}. In fact, adding the label information only allow a modest improvement of our method, and is outperformed in particular by the deep graph kernels, WL and PSCN, see Table \ref{bio1-3}.

\section{Human brains dataset}\label{humanbds}
In this section we  apply our method on a particular brains dataset we build from magnetic resonance imaging (MRI) data. Details about its creation can be found in the Appendix (\ref{apendix}). 

\subsection{Dataset}
We introduce a new human brains dataset (\textbf{BRAINS}) representing them as undirected networks (connectomes). Structural and diffusion MRI data of 91 healthy men and 113 healthy woman is preprocessed in order to create final connectomes. All graphs have the same 84 nodes representing neural Regions of Interests (ROIs). Weighted edges correspond to the number of neural fibers linking two ROIs. Node-label encodes the ROI correspondence. The task is to classify connectomes as male or female. Data is provided by email request to the authors.

\subsection{Experiments and results}
The aim of this experiment is to show how our method can be smoothly adapted when  applied to a particular dataset in which all graphs take place on the same set of vertices and labels. The experimental setting is the same that previous section but using  linear C-SVM as classifier. We compare our method with the algorithms performing well in previous experiments and that are able to exploit nodal information. We report \textit{average prediction accuracies} and \textit{standard deviations}.

\subsubsection{Vertex labeling trick} 
Exploiting node label data as we did in the bioinformatic benchmark is useless in this case. Because ROIs, in bijection with vertices,  are the same along the equal-sized graphs, it would generate redundant features. Instead, we one-hot encode each vertex-label (generating the characteristic binary vector of each ROI) and then computing the generalized assortativities (\ref{vector_covariance}).
\subsubsection{Results}
As can bee seen in Table \ref{brains}, even ignoring the node labels (ROIs) our baseline features (DyF-nolab) outperform all WL, DWL, SP and DSP kernels, and feature-based approach. Graph-kernels are not able to exploit efficiently nodal data in the case all graphs contain the same set of vertex labeling assignment. Including edge weights in our formulation (\ref{markov}) is straightforward and also improves the accuracy of our method (DyF-nolab + w). Finally, including node labels through the one-hot encoding trick shows a considerably improvement over all approaches (DyF + w).

\section*{Discussion}
We devised a new feature-based method to discriminate between different classes of networks. We define generalized assortativities features by setting up a dynamics on the network and computing covariances over divers network attributes among multiples time scales. It turns out that those features on a small number of structural features, in addition to node metadata whenever present are useful to characterize networks and classification proposes, achieving high accuracies on (social and biochemical) datasets without node metadata, and fair accuracy on biochemical datasets with node metadata. It may be thought that the assortativity patterns of node labels (atom type, protein label, etc) do not present a significant variation, therefore requiring a more in-depth exploration of the meta-data structure as offered by some competing methods, while the structural attribute are easily captured by their assortativities.

A natural extension of our work would be considering features assessed by different types of dynamics, including different types of discrete-time or continuous time random walks taking place on the network \cite{Lambiotte2014}. We could potentially exploit the prior knowledge about the flow pattern or the dynamics we consider and then tailoring the stability measure in order to reveal more relevant structural attributes for specific problems and potentially reaching better accuracies.

\begin{table}[!t]
\scriptsize
\caption{\textit{Human brains} Mean and standard deviation of classification accuracy for Weisfeiler-Lehman (WL) \cite{Shervashidze:2011:WGK:1953048.2078187}, Deep Weisfeiler-Lehman (DWL) \cite{Yanardag:2015:DGK:2783258.2783417}, Shortest-path (SP) \cite{Borgwardt}, Deep Shortest-path (DSP) \cite{Yanardag:2015:DGK:2783258.2783417} kernels, Feature-Based (FB) \cite{Barnett2016}, Dynamics Based Features (DyF) (our method) }\label{brains}
\begin{center}
\begin{tabular}{l@{\quad}c@{\quad}c@{\quad}c@{\quad}c@{\quad}}
\hline  \rule{0pt}{12pt} 
Dataset	&	WL	&	DWL	&	SP	&	DSP	\\
\hline  \rule{0pt}{12pt} 
BRAINS	&	61.20 $ \pm $ 2.16	&	59.55 $ \pm $ 2.11	&	65.45 $ \pm $ 1.78	&	65.70 $ \pm $ 1.94	\\
\hline \rule{0pt}{12pt} 
\end{tabular}
\scriptsize
\begin{tabular}{l@{\quad}c@{\quad}c@{\quad}c@{\quad}c@{\quad}}
\hline  \rule{0pt}{12pt} 
Dataset	&	FB	&	DyF-nolab	&	DyF-nolab + w	&	DyF + w	\\
\hline  \rule{0pt}{12pt} 
BRAINS	&	65.95 $ \pm $ 2.54	&	\textbf{67.75 $ \pm $ 1.25}	&	\textbf{75.05 $ \pm $ 0.98}	&	\textbf{81.30 $ \pm $ 2.07}	\\
\hline \rule{0pt}{12pt} 
\end{tabular}
\end{center}
\end{table}
\subsubsection{Acknowledgments}
The authors acknowledges support from the grant ``Actions de recherche concertées (ARC) and IAP DYSCO (Dynamical Systems Control and Optimization) of the Communauté Française de Belgique. We also thank Marco Saerens, Leto Peel and Roberto D'Ambrosio for helpful discussions and suggestions.

Data were provided by the Human Connectome Project, WU-Minn Consortium (Principal Investigators: David Van Essen and Kamil Ugurbil; 1U54MH091657) funded by the 16 NIH Institutes and Centers that support the NIH Blueprint for Neuroscience Research; and by the McDonnell Center for Systems Neuroscience at Washington University.
%
%
\bibliographystyle{plain}

\bibliography{graph_class}
\section{Appendix: Human Brains Dataset}\label{apendix}
Representing the human brain as a complex network has gained much interest in the neuroscientific community over the last years \cite{fornito2016fundamentals}. Indeed, vertices and edges are natural analogies of neurons and axons. However, mapping the billions of neurons and trillions of axons composing the human brain remains impossible. By defining some Regions Of Interest (ROIs) over the cerebral cortex and measuring the relationship between them, macroscale connectomics aims at studying the connectivity of the human brain thanks to graph theoretical tools.\\
Depending on the kind of study to conduct, structural or functional connectomes (i.e., brain networks) can be defined. The former most often define anatomical links between two ROIs by evaluating the number of shared nerve fibers, while the latter measure the amplitude of the co-activity between two ROIs during some tasks. Intuitively, structural connectivity can be seen as the substrate from which emerge co-activity patterns, and somehow dominates functional connectivity.
\subsection{Methods}
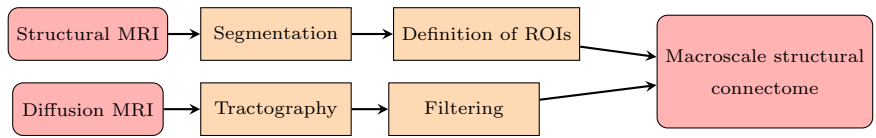
\begin{figure}[!ht]
\centering
\begin{tikzpicture}[node distance=2cm]
\node[draw] at (-7,1) (structMRI) [startstop] {\scriptsize{Structural MRI}};
\node[draw] at (-7,0) (diffMRI) [startstop] {\scriptsize{Diffusion MRI}};
\node[draw] at (-4.5,1) (seg) [process] {\scriptsize{Segmentation}};
\node[draw] at (-1.7,1) (roi) [process] {\scriptsize{Definition of ROIs}};
\node[draw] at (-4.5,0) (tract) [process] {\scriptsize{Tractography}};
\node[draw] at (-2,0) (filtering) [process] {\scriptsize{Filtering}};
\node[draw, align=center] at (2,0.5) (connectome) [stop] {\scriptsize{Macroscale structural}\\\scriptsize{connectome}};
\draw [arrow] (structMRI) -- (seg);
\draw [arrow] (seg) -- (roi);
\draw [arrow] (diffMRI) -- (tract);
\draw [arrow] (roi) -- (connectome);
\draw [arrow] (tract) -- (filtering);
\draw [arrow] (filtering) -- (connectome);
\end{tikzpicture}
\caption{Processing pipeline used for structural connectome generation.}
\label{fig:flowchart_connectome}
\end{figure}

To build a structural connectome, informations from both structural and diffusion MRI must be combined. While structural MRI will help to define the ROIs, i.e. the vertices of the connectome, diffusion MRI will allow to reconstruct the nerve fibers of the white matter and to define the weights of the edges of the resulting brain network. Figure \ref{fig:flowchart_connectome} and the following sections present the main steps of a widespread method \cite{fornito2016fundamentals,hagmann2007mapping} allowing to build a macroscale structural connectome.
Preprocessed structural and diffusion MRI data of 91 healthy men and 113 healthy women, aged from 22 to 37 years, have been retrieved from the Human Connectome Project database. The whole processing (including registration of both images in order to get voxel correspondence) was performed by using the MRtrix3 toolbox. 
\subsubsection{Strucural MRI processing}
T1-weighted MRI images are used to provide good contrast between white matter and gray matter. The former consists of the inner part of the brain and is mainly composed of axons (also called tracts), whereas the latter contains the neuronal cells. Hence, image segmentation  was first performed to separate white and gray matter in order to build anatomical information that will constrain further processing steps.\\
Next, 34 cortical ROIs per hemisphere were defined according to the Desikan-Killiany automated atlas \cite{desikan2006automated}, and 8 subcortical ROIs per hemisphere were segmented following the procedure described in \cite{patenaude2011bayesian}. It is interesting to notice that, following this procedure, the whole set of resulting connectomes will share the same 84 vertices.
\subsubsection{Diffusion MRI processing and tractography}
Diffusion imaging allows to identify pathways inside the cerebral cortex by measuring the diffusivity of water molecules inside the brain. In white matter, this diffusivity is highly constrained by the direction of nerve fibers. Therefore, the measurement of the diffusivity along several directions allows to deduce a privileged direction at a given voxel of the image. In fact, the voxel size in diffusion imaging ranges from 1 to 2 $mm$, while axons have diameter of approximatively 1 $\mu m$. Thus, the reconstructed pathways correspond more to axonal beams than to single fibers. However, at the macroscale, these axonal beams provide sufficient information about the connectivity between ROIs.

Tractography is the process allowing to reconstruct axonal beams by following the privileged direction of diffusivity from voxels to voxels. Probabilistic rather than deterministic algorithms are often chosen since they allow to reduce reconstruction errors by evaluating, at each step, the probability density of a connection between two voxels. The whole-brain anatomically-constrained tractography \cite{smith2012anatomically} was thus performed by combining an improved probabilistic algorithm based on a second-order integration over the fibers orientation probability density \cite{tournier2010improved} and previous anatomical information. Spherical-deconvolution filtering of reconstructed axonal beams \cite{smith2013sift} was then applied to remove potential biases and get even more biologically meaningful results.
\subsubsection{Connectome generation}
\begin{figure}[!th]
\begin{center}
\includegraphics[scale=0.13]{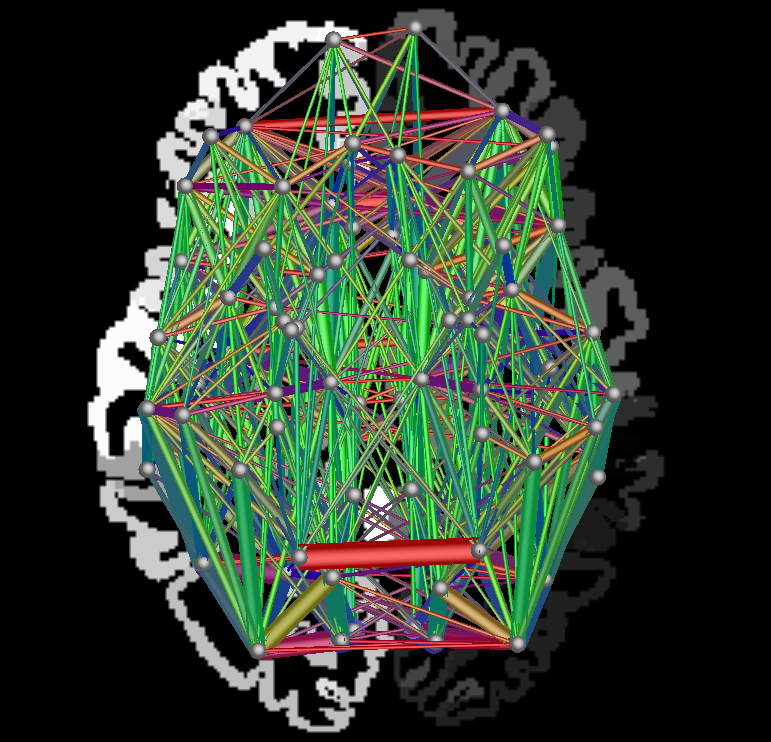}
\includegraphics[scale=0.13]{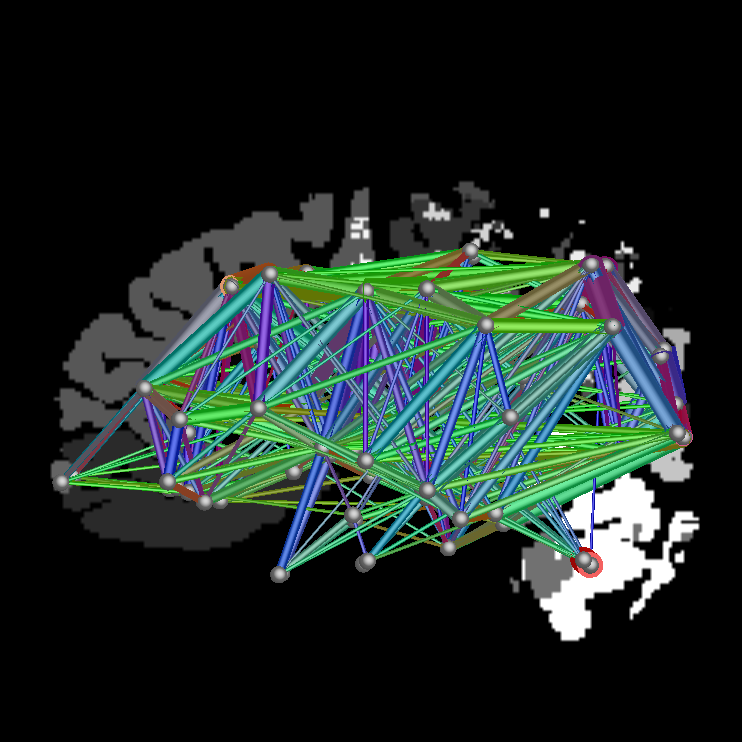}
\caption{Structural connectome: abstract representation of the structure of the human brain as a complex network. Gray spheres represent the ROIs. The color of the edges corresponds to the orientation. Left: axial view. Right: sagittal view.}
\label{fig:connectome}
\end{center}
\end{figure}

The final step involves counting the reconstructed fibers linking any two ROIs to define the weight of the corresponding edge. Hence, the generated networks are weighted-undirected, as depicted in Figure \ref{fig:connectome}. Furthermore, as already reported in the literature, these connectomes exhibit topological properties such as small-worldness, modularity and rich-club organization for instance, which confirms the validity of the generation process.

\end{document}